\documentclass[sigconf]{acmart}
\pdfoutput=1

\usepackage{balance}
\usepackage{threeparttable}
\usepackage{graphicx}
\usepackage{subfigure}
\usepackage{cases}
\usepackage{tabularx}

\def\BibTeX{{\rm B\kern-.05em{\sc i\kern-.025em b}\kern-.08emT\kern-.1667em\lower.7ex\hbox{E}\kern-.125emX}}
    
\copyrightyear{2019} 
\acmYear{2019} 
\setcopyright{acmcopyright}
\acmConference[KDD '19]{The 25th ACM SIGKDD Conference on Knowledge Discovery and Data Mining}{August 4--8, 2019}{Anchorage, AK, USA}
\acmBooktitle{The 25th ACM SIGKDD Conference on Knowledge Discovery and Data Mining (KDD '19), August 4--8, 2019, Anchorage, AK, USA}
\acmPrice{15.00}
\acmDOI{10.1145/3292500.3330680}
\acmISBN{978-1-4503-6201-6/19/08}

\begin{document}

\title{Time-Series Anomaly Detection Service at Microsoft}

\author{Hansheng Ren, Bixiong Xu, Yujing Wang, Chao Yi, Congrui Huang, Xiaoyu Kou} \authornote{Hansheng Ren is a student in University of Chinese Academy of Sciences; Chao Yi and Xiaoyu Kou are students in Peking University. The work was done when they worked as full-time interns at Microsoft.}
  \author{Tony Xing, Mao Yang, Jie Tong, Qi Zhang}
    \affiliation{ 
      \institution{Microsoft}
      \state{Beijing}
      \country{China}
    }
    \email{{v-hanren,bix,yujwang,t-chyi,conhua,v-xiko,tonyxin,maoyang,jietong,qizhang}@microsoft.com}

%
\renewcommand{\shortauthors}{Ren and Xu, et al.}

%
\begin{abstract}
Large companies need to monitor various metrics (for example, Page Views and Revenue) of their applications and services in real time. At Microsoft, we develop a time-series anomaly detection service which helps customers to monitor the time-series continuously and alert for potential incidents on time. In this paper, we introduce the pipeline and algorithm of our anomaly detection service, which is designed to be accurate, efficient and general. The pipeline consists of three major modules, including data ingestion, experimentation platform and online compute. To tackle the problem of time-series anomaly detection, we propose a novel algorithm based on Spectral Residual (SR) and Convolutional Neural Network (CNN). Our work is the first attempt to borrow the SR model from visual saliency detection domain to time-series anomaly detection. Moreover, we innovatively combine SR and CNN together to improve the performance of SR model. Our approach achieves superior experimental results compared with state-of-the-art baselines on both public datasets and Microsoft production data.
\end{abstract}

%
%


%

\maketitle

\section{Introduction}
Anomaly detection aims to discover unexpected events or rare items in data. It is popular in many industrial applications and is an important research area in data mining. Accurate anomaly detection can trigger prompt troubleshooting, help to avoid loss in revenue, and maintain the reputation and branding for a company. For this purpose, large companies have built their own anomaly detection services to monitor their business, product and service health~\cite{yahooo2015EGADS,twitter-ad}. When anomalies are detected, alerts will be sent to the operators to make timely decisions related to incidents. For instance, Yahoo releases EGADS~\cite{yahooo2015EGADS} to automatically monitor and raise alerts on millions of time-series of different Yahoo properties for various use-cases. At Microsoft, we build an anomaly detection service to monitor millions of metrics coming from Bing, Office and Azure, which enables engineers move faster in solving live site issues. In this paper, we focus on the pipeline and algorithm of our anomaly detection service specialized for time-series data. 

There are many challenges in designing an industrial service for time-series anomaly detection: 

\textbf{Challenge 1: Lack of Labels.} To provide anomaly detection services for a single business scenario, the system must process millions of time-series simultaneously. There is no easy way for users to label each time-series manually. Moreover, the data distribution of time-series is constantly changing, which requires the system recognizing the anomalies even though similar patterns have not appeared before. That makes the supervised models insufficient in the industrial scenario. 

\begin{figure}[htbp]
	\centering 
	\subfigure[seasonal]{
		\label{Fig.seasonal}
		\includegraphics[width=0.14\textwidth]{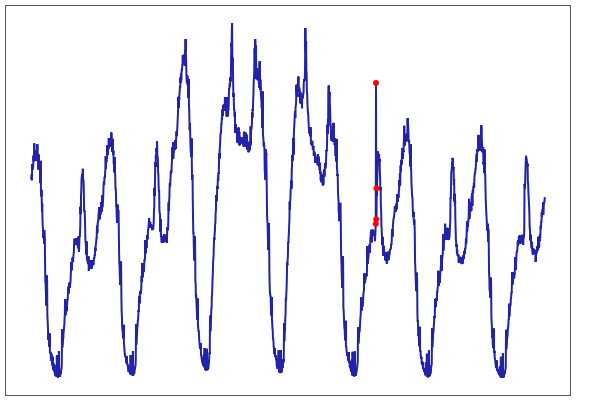}}
	\subfigure[stable]{
		\label{Fig.stable}
		\includegraphics[width=0.14\textwidth]{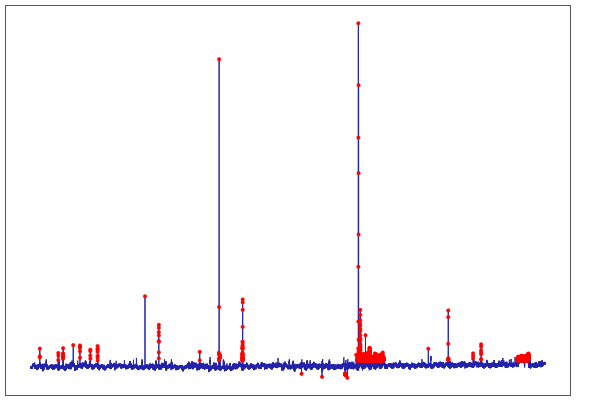}}
	\subfigure[unstable]{
		\label{Fig.unstable}
		\includegraphics[width=0.14\textwidth]{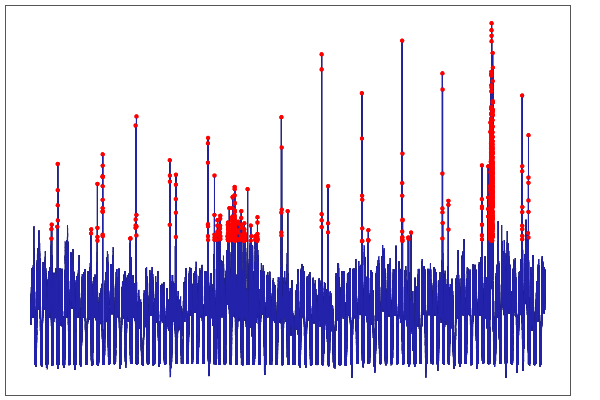}}
	\caption{Different types of time-series.}
	\label{fig:series_pattern}
\end{figure}

\textbf{Challenge 2: Generalization.} Various kinds of time-series from different business scenarios are required to be monitored. As shown in Figure \ref{fig:series_pattern}, there are several typical categories of time-series patterns; and it is important for industrial anomaly detection services to work well on all kinds of patterns. However, existing approaches are not generalized enough for different patterns. For example, Holt winters~\cite{holtwinters} always shows poor results in (b) and (c); and Spot~\cite{siffer2017anomaly} always shows poor results in (a). Thus, we need to find a solution of better generality. 

\textbf{Challenge 3: Efficiency.} In business applications, a monitoring system must process millions, even billions of time-series in near real time. Especially for minute-level time-series, the anomaly detection procedure needs to be finished within limited time. Therefore, efficiency is one of the major prerequisites for online anomaly detection service. Even though the models with large time complexity are good at accuracy, they are often of little use in an online scenario.

To tackle the aforementioned problems, our goal is to develop an anomaly detection approach which is accurate, efficient and general. Traditional statistical models~\cite{twitter-ad,siffer2017anomaly,holtwinters,rosner1983percentage,lu2009network,mahimkar2011rapid,zhang2005network,rasheed2009fourier} can be easily adopted online, but their accuracies are not sufficient for industrial applications. Supervised models~\cite{liu2015opprentice,anomalyGoogle} are superior in accuracy, but they are insufficient in our scenario because of lacking labeled data. There are other unsupervised approaches, for instance, Luminol~\cite{linkedin/luminol} and DONUT~\cite{xu2018unsupervised}. However, these methods are either too time-consuming or parameter-sensitive. Therefore, we aim to develop a more competitive method in the unsupervised manner which favors accuracy, efficiency and generality simultaneously. 

In this paper, we borrow the Spectral Residual model \cite{hou2007saliency} from the visual saliency detection domain to our anomaly detection application. Spectral Residual (SR) is an efficient unsupervised algorithm, which demonstrates outstanding performance and robustness in the visual saliency detection tasks. To the best of our knowledge, our work is the first attempt to borrow this idea for time-series anomaly detection. The motivation is that the time-series anomaly detection task is similar to the problem of visual saliency detection essentially. Saliency is what "stands out" in a photo or scene, enabling our eye-brain connection to quickly (and essentially unconsciously) focus on the most important regions. Meanwhile, when anomalies appear in time-series curves, they are always the most salient part in vision.

Moreover, we propose a novel approach based on the combination of SR and CNN. CNN is a state-of-the-art method for supervised saliency detection when sufficient labeled data is available; while SR is a state-of-the-art approach in unsupervised setting. Our innovation is to unite these two models by applying CNN on the basis of SR output directly. As the problem of anomaly discrimination becomes much easier upon the output of SR model, we can train CNN through automatically generated anomalies and achieve significant performance enhancement over the original SR model. Because the anomalies used for CNN training is fully synthetic, the SR-CNN approach remains unsupervised and establishes a new state-of-the-art performance when no manually labeled data is available.

As shown in the experiments, our proposed algorithm is more accurate and general than state-of-the-art unsupervised models. Furthermore, we also apply it as an additional feature in the supervised learning model. The experimental results demonstrate that the performance can be further improved when labeled data is available; and the additional features do provide complementary information to existing anomaly detectors. Up to the date of paper submission, the $F_1$-score of our unsupervised and supervised approaches are both the best ever achieved on the open datasets. 

The \textbf{contributions} of this paper are highlighted as below:
\begin{itemize}
  \item For the first time in the anomaly detection field, we borrow the technique of visual saliency detection to detect anomalies in time-series data. The inspiring results prove the possibility of using computer vision technologies to solve anomaly detection problems.
  \item We combine the SR and CNN model to improve the accuracy of time-series anomaly detection. The idea is innovative and the approach outperforms current state-of-the-art methods by a large margin. Especially, the $F_1$-score is improved by more than 20\% on Microsoft production data. 
  \item From the practical perspective, the proposed solution has good generality and efficiency. It can be easily integrated with online monitoring systems to provide quick alerts for important online metrics. This technique has enabled product teams to move faster in detecting issues, save manual efforts, and accelerate the process of diagnostics.
  \end{itemize}
  
The rest of this paper is organized as follows. First, in Section \ref{sys_design}, we describe the details of system design, including data ingestion, experimentation platform and online compute. Then, we share our experience of real applications in Section 3 and introduce the methodology in Section 4. Experimental results are analyzed in Section 5 and related works are presented in Section 6. Finally, we conclude our work and put forward future work in Section 7. 
\section{system overview}
\label{sys_design}
\begin{figure*}[t]
    \centering
    \includegraphics[width=\textwidth]{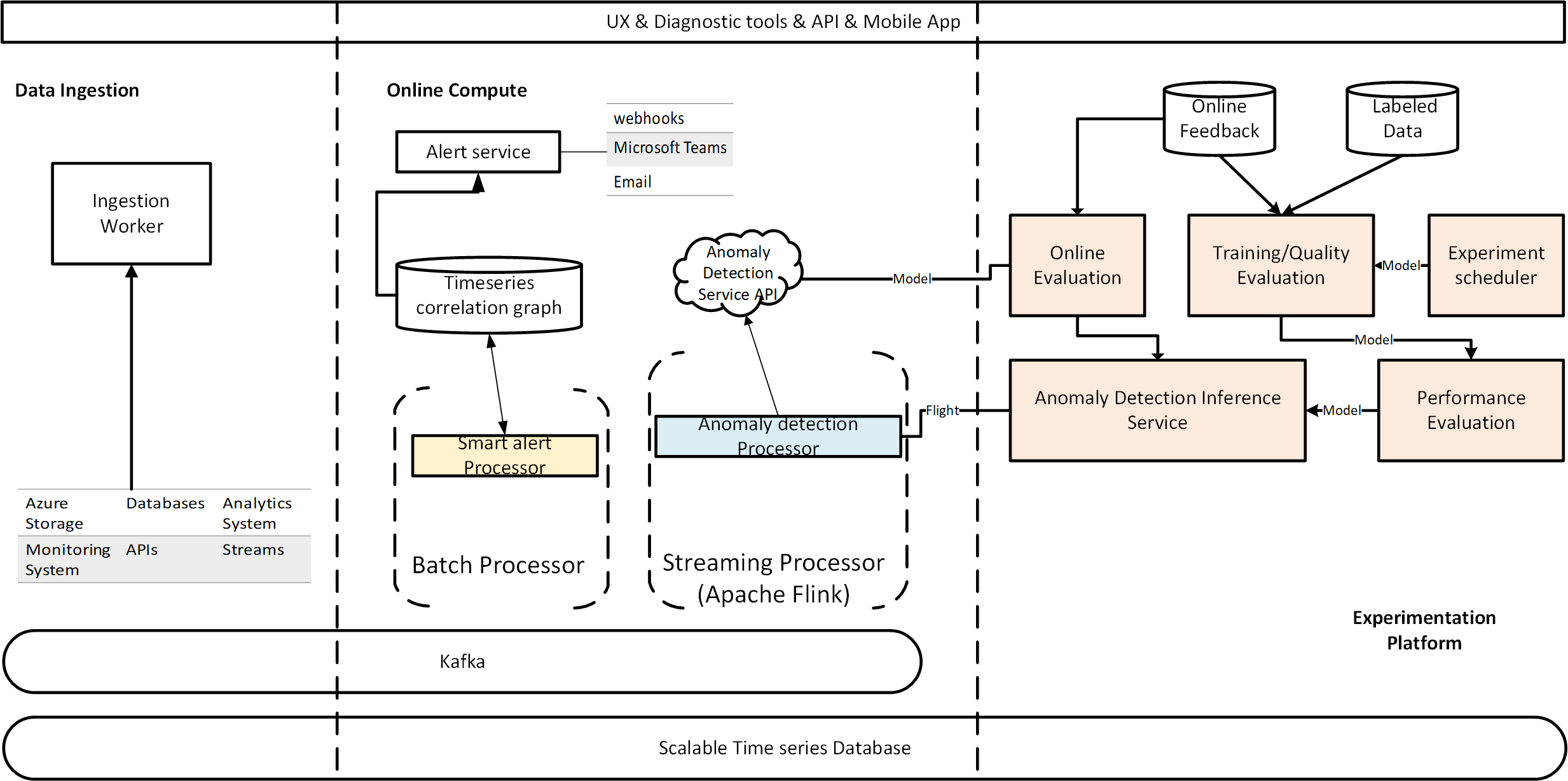}
    \caption{System Overview}
\end{figure*}
The whole system consists of three major components: \textbf{data ingestion}, \textbf{experimentation platform} and \textbf{online compute}. Before going into more detail about these components, we will introduce the whole pipeline first. Users can register monitoring tasks by ingesting time-series to the system. Ingesting time-series from different data sources (including Azure storage, databases and online streaming data) is supported. The \textit{ingestion worker} is responsible for updating each time-series according to the designated granularity, for example, minute, hour, or day. Time-series points enter the streaming pipeline through Kafka and is stored into the time-series database. \textit{Anomaly detection processor} calculates the anomaly status for incoming time-series points online. In a common scenario of monitoring business metrics, users ingest a collection of time-series simultaneously. As an example, Bing team ingests the time-series representing the the usage of different markets and platforms. When incident happens, \textit{alert service} combines anomalies of related time-series and sends them to users through emails and paging services. The combined anomalies show the overall status of an incident and help users to shorten the time in diagnosing issues. Figure 2 illustrates the general pipeline of the system.
\subsection{Data Ingestion} 
Users can register a monitor task by creating a \textit{Datafeed}. Each datafeed is identified by \textit{Connect String} and \textit{Granularity}. Connect String is used to connect user's storage system to the anomaly detection service. Granularity indicates the update frequency of a datafeed; and the minimum granularity is one minute. An ingestion task will ingest the data points of time-series to the system according to the given granularity. For example, if a user sets minute as the granularity, ingestion module will create a task every minute to ingest a new data point. Time-series points are ingested into influxDB\footnote{https://www.influxdata.com/} and Kafka\footnote{https://kafka.apache.org/}. Throughput of this module varies from 10,000 to 100,000 data points per second.
\subsection{Online Compute}
The online compute module processes each data point immediately after it enters the pipeline. To detect anomaly status of an incoming point, a sliding window of the time-series data points is required. Therefore, we use Flink\footnote{https://flink.apache.org/} to manage the points in memory to optimize the computation efficiency. Currently, the streaming pipeline processes more than 4 million time-series every day in production. The maximum throughput can be 4 million every minute. \textit{Anomaly detection processor} detects anomalies for each single time-series. In practice, a single anomaly is not enough for users to diagnose their service efficiently. Thus, \textit{smart alert processor} correlates the anomalies from difference time-series and generates an incident report accordingly. As anomaly detection is the main topic in this paper, smart alert is not discussed in more detail.
\subsection{Experimentation Platform} 
We build an experimentation platform to evaluate the performance of anomaly detection models. Before we deploy a new model, offline experiments and online A/B tests will be conducted on the platform. Users can mark a point as anomaly or not on the portal. A labeling service is provided to human editors. Editors will first label true anomaly points of a single time-series and then label false anomaly points from anomaly detection results of a specific model. Labeled data is used to evaluate the accuracy of the anomaly detection model. We also evaluate the efficiency and generality of each model on the platform. In online experiments, we flight several datafeeds to the new model. A couple of metrics, such as click through rate of alerts, percentage of anomalies and false anomaly rate is used to decide whether the new model can be deployed to production. The experimentation platform is built on Azure machine learning service\footnote{https://azure.microsoft.com/en-us/services/machine-learning-service/}. If a model is verified to be effective, the platform will expose it as a web service and host it on K8s\footnote{https://kubernetes.io/docs/concepts/overview/what-is-kubernetes/}.
\section{applications}
\begin{figure*}[t]
    \subfigure[Alert Page]{
        \label{fig:alert}
		\includegraphics[width=0.45\textwidth,height=5cm]{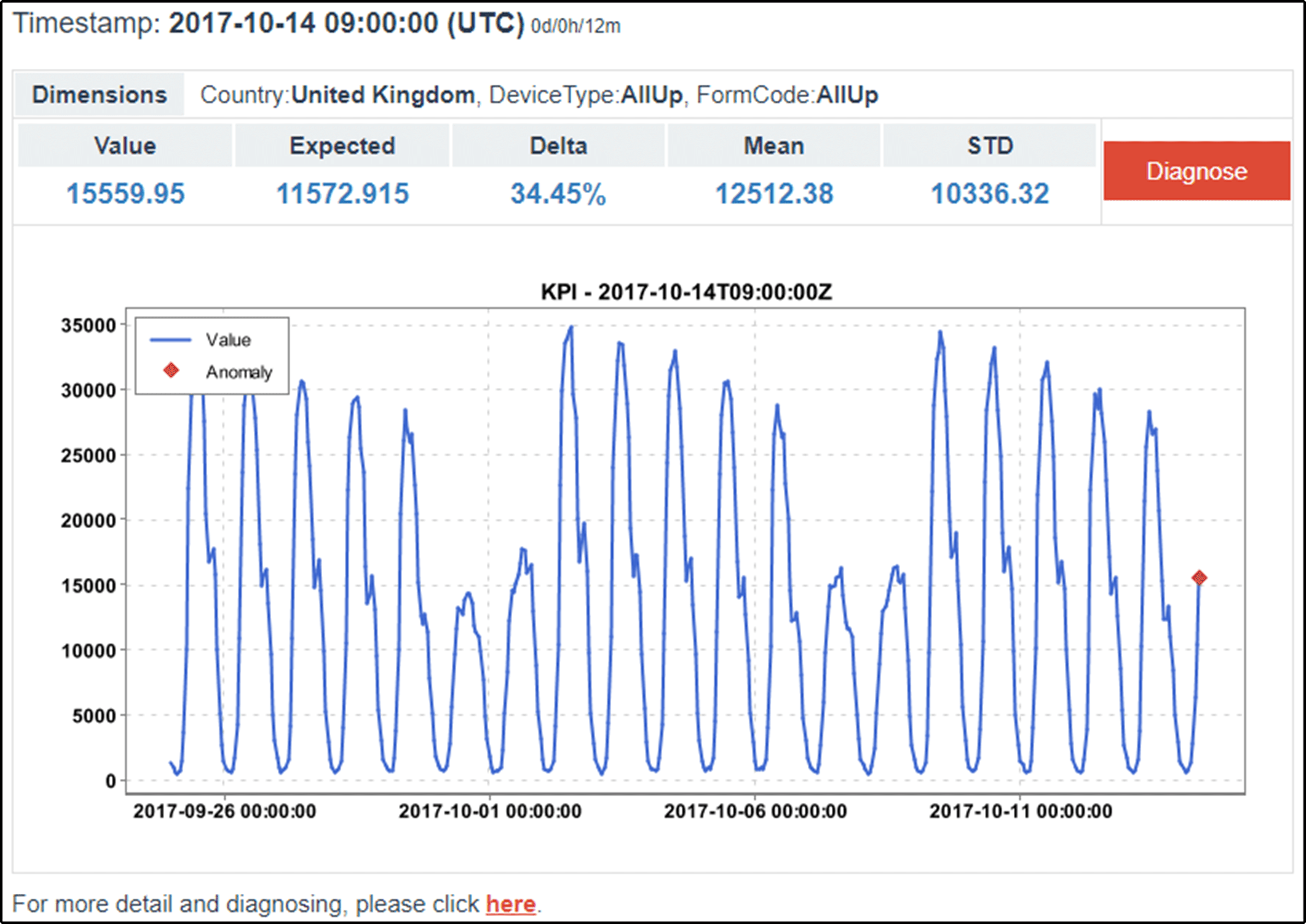}}
		\hspace*{\fill}
	 \subfigure[Incident Report]{
		 \label{fig:incident}
	    \includegraphics[width=0.45\textwidth,height=5cm]{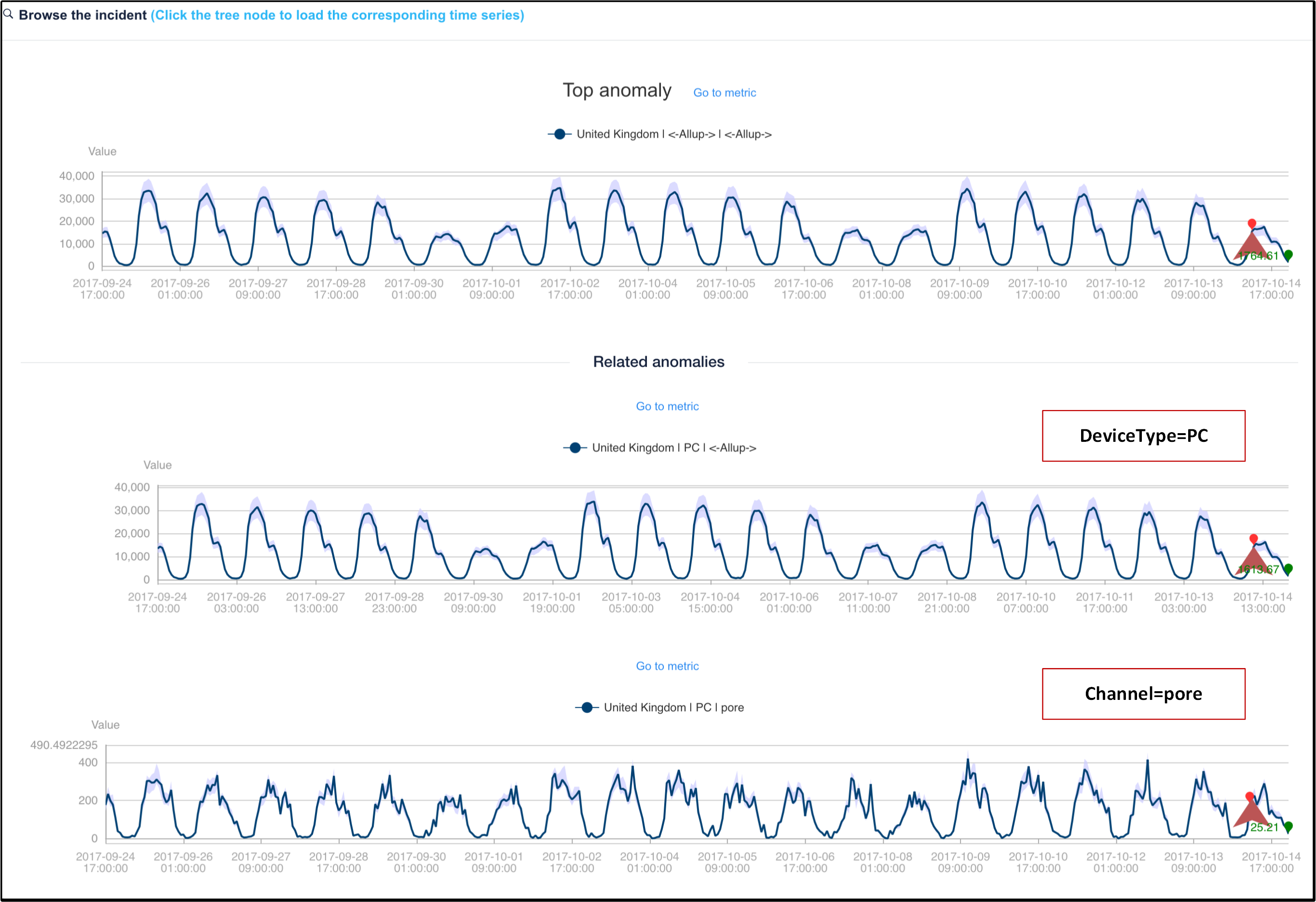}}
	 \caption{An illustration of example application from Microsoft Bing}
	 \label{fig:sys_exp}
\end{figure*}
At Microsoft, it is a common need to monitor business metrics and act quickly to address the issue if there is anything outside of the normal pattern. To tackle the problem, we build a scalable system with the ability to monitor minute-level time-series from various data sources. Automated diagnostic insights are provided to assist users to resolve their issues efficiently. The service has been used by more than 200 product teams within Microsoft, across Office 365, Windows, Bing and Azure organizations, with more than 4 million time-series ingested and monitored continuously.
	
As an example, Michael from Bing team would like to monitor the usage of their service in the global marketplace. In the anomaly detection system, he created a new \textit{datafeed} to ingest thousands of time-series, each indicating the usage of a specific market (US, UK, etc.), device (PC, windows phone, etc.) or channel (PORE, QBRE, etc.). Within 5 minutes, Michael saw the ingested time-series on the portal. At 9am, Oct-14, 2017, the time-series associated to the UK market encountered an incident. Michael was notified through E-mail alerts (as shown in Figure \ref{fig:alert}) and started to investigate the problem. He opened the incident report where the top correlated time-series with anomalies are selected from a set of time-series around 9am. As shown in Figure \ref{fig:incident}, usage on PC devices and PORE channel can be found in the incident report. Michael brought this insight to the team and finally found that the problem was caused by a relevance issue which made users do lots of pagination requests (PORE) to get satisfactory search results.
    
As another example, the Outlook anti-spam team used to leverage a rule-based method to monitor the effectiveness of their spam detection system. However, this method was not easy to be maintained and usually showed bad cases on some Geo-locations. Therefore, they ingested key metrics to our anomaly detection service to monitor the effectiveness of their spam detection model across different Geo-locations. Through our API, they have integrated anomaly detection ability into the Office DevOps platform. By using this automatic detection service, they have covered more Geo-locations and received less false positive cases compared to the original rule-based solution.
\section{methodology}
The problem of time-series anomaly detection is defined as below. 
\newtheorem{problem}{Problem}
\begin{problem}
\label{def}
	Given a sequence of real values, i.e., $\textbf{x}={x_1, x_2, ..., x_n}$, the task of time-series anomaly detection is to produce an output sequence, $\textbf{y}={y_1, y_2, ..., y_n}$, where $y_i \in \{0, 1\}$ denotes whether $x_i$ is an anomaly point.
\end{problem}
As emphasized in the Introduction, our challenge is to develop a general and efficient algorithm with no labeled data. Inspired by the domain of visual computing, we adopt Spectral Residual (SR)~\cite{hou2007saliency}, a simple yet powerful approach based on Fast Fourier Transform (FFT)~\cite{van1992computational}. The SR approach is unsupervised and has been proved to be efficient and effective in visual saliency detection applications. We believe that the visual saliency detection and time-series anomaly detection tasks are similar essentially, because the anomaly points are usually salient in the visual perspective. 
	
Furthermore, recent saliency detection research has shown favor to end-to-end training with Convolutional Neural Networks (CNNs) when sufficient labeled data is available~\cite{zhao2015saliency}. Nevertheless, it is prohibitive for our application as large-scale labeled data is difficult to be collected online. As a trade-off, we propose a novel method, SR-CNN, which applies CNN on the output of SR model directly. CNN is responsible to learn a discriminate rule to replace the single threshold adopted by the original SR solution. The problem becomes much easier to learn the CNN model on SR results than on the original input sequence. Specifically, we can use artificially generated anomaly labels to train the CNN-based discriminator. In the following sub-sections, we introduce the details of SR and SR-CNN methods respectively. 
\subsection{SR (Spectral Residual)}
\label{section:spectral residual}
The Spectral Residual (SR) algorithm consists of three major steps: (1) Fourier Transform to get the log amplitude spectrum; (2) calculation of \textit{spectral residual}; and (3) Inverse Fourier Transform that transforms the sequence back to spatial domain. Mathematically, given a sequence $\textbf{x}$, we have
\begin{align}
& A(f) = Amplitude(\mathfrak{F}(\textbf{x})) \\
& P(f) = Phrase(\mathfrak{F}(\textbf{x})) \\
& L(f) = log(A(f)) \\
& AL(f) = h_q(f) \cdot L(f) \\
& R(f) = L(f) - AL(f) \\
& S(\textbf{x}) = \left\Vert \mathfrak{F}^{-1}(exp(R(f) + iP(f))) \right\Vert
\end{align}
where $\mathfrak{F}$ and $\mathfrak{F}^{-1}$ denote Fourier Transform and Inverse Fourier Transform respectively. \textbf{x} is the input sequence with shape $n \times 1$; $A(f)$ is the amplitude spectrum of sequence \textbf{x}; $P(f)$ is the corresponding phase spectrum of sequence \textbf{x}; $L(f)$ is the log representation of $A(f)$; and $AL(f)$ is the average spectrum of $L(f)$ which can be approximated by convoluting the input sequence by $h_q(f)$, where $h_q(f)$ is an $q \times q$ matrix defined as: 
\begin{align*}
h_q(f)
=
\frac{1}{q^{2}}
\begin{bmatrix}
1 & 1 & 1 & \dots  & 1 \\
1 & 1 & 1 & \dots  & 1 \\
\vdots & \vdots & \vdots & \ddots & \vdots \\
1 & 1 & 1 & \dots  & 1
\end{bmatrix}
\end{align*}
    
$R(f)$ is the \textit{spectral residual}, i.e., the log spectrum $L(f)$ subtracting the averaged log spectrum $AL(f)$. The \textit{spectral residual} serves as a compressed representation of the sequence while the innovation part of the original sequence becomes more significant. At last, we transfer the sequence back to spatial domain via Inverse Fourier Transform. The result sequence $S(\textbf{x})$ is called the \textit{saliency map}.
    
\begin{figure}[t]
\centering
\includegraphics[width=0.45\textwidth]{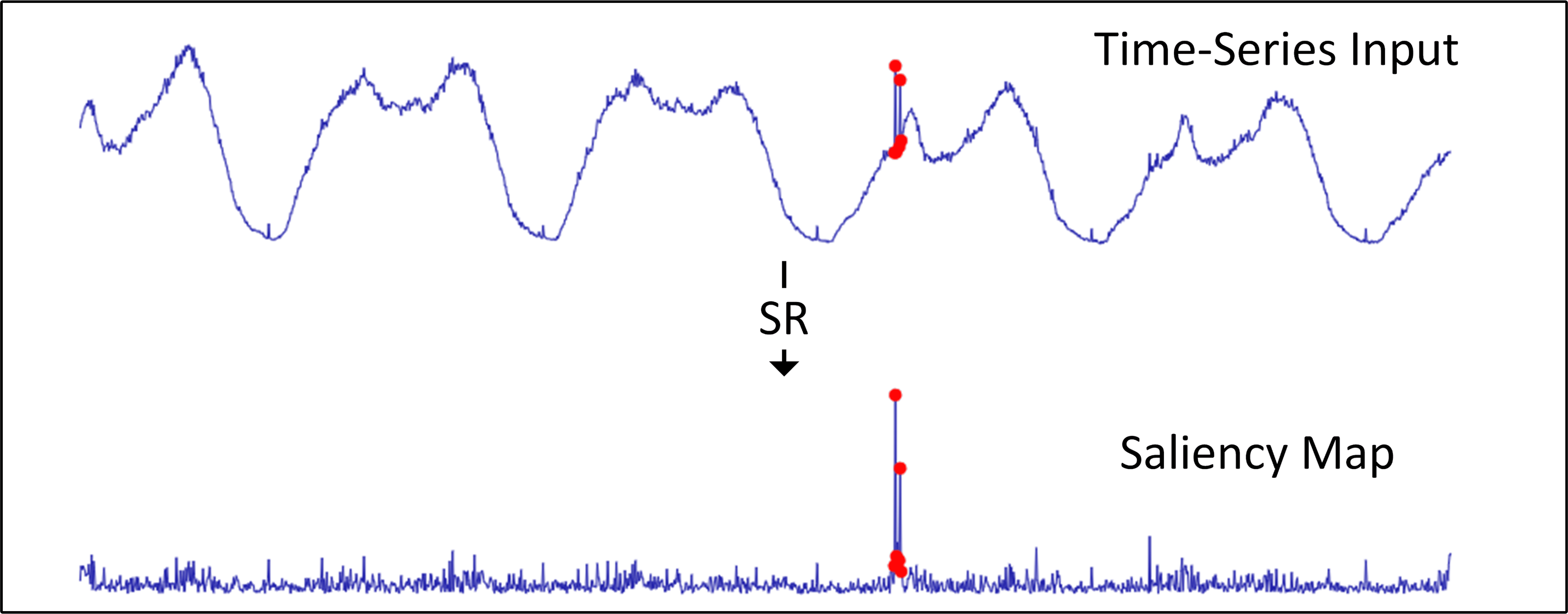}
\caption{Example of SR model results}
\label{fig:sr_example}
\end{figure}
    
Figure \ref{fig:sr_example} shows an example of the original time-series and the corresponding \textit{saliency map} after SR processing. As shown in the figure, the innovation point (shown in red) in the \textit{saliency map} is much more significant than that in the original input. Based on the \textit{saliency map}, it is easy to leverage a simple rule to annotate the anomaly points correctly. We adopt a simple threshold $\tau $ to annote anomaly points. Given the saliency map $S(\textbf{x})$, the output sequence $O(\textbf{x})$ is computed by:
    
\begin{align}
O(x_i) = 
\begin{cases}
1, &\text{if $\frac{S(x_i) - \overline{S(x_i)}}{\overline{S(x_i)}}) > \tau $,} \\
	0, &\text{otherwise,}
\end{cases}
\label{fuc:threshold}
\end{align}
where $x_i$ represents an arbitrary point in sequence \textbf{x}; $S(x_i)$ is the corresponding point in the saliency map; and $\overline{S(x_i)}$ is the local average of the preceding z points of $S(x_i)$.

In practice, the FFT operation is conducted within a \textit{sliding window} of the sequence. Moreover, we expect the algorithm to discover the anomaly points with low latency. That is, given a stream $x_1, x_2, ..., x_n$ where $x_n$ is the recent point, we want to tell if $x_n$ is an anomaly point as soon as possible. However, the SR method works better if the target point locates in the center of the sliding window. Thus, we add several \textit{estimated points} after $x_n$ before inputting the sequence to SR model. The value of estimated point $x_{n+1}$ is calculated by: 
\begin{align}
\overline{g} = \frac{1}{m}\sum_{i=1}^{m}{g(x_{n}, x_{n-i})} \\
	x_{n+1} = x_{n-m+1} + \overline{g} \cdot m
\end{align}
where $g(x_i, x_j)$ denotes the gradient of the straight line between point $x_i$ and $x_j$; and $\overline{g}$ represents the average gradient of the preceding points. $m$ is the number of preceding points considered, and we set $m=5$ in our implementation. We find that the first estimated point plays a decisive role. Thus, we just copy $x_{n+1}$ for $\kappa$ times and add the points to the tail of the sequence.
	
To summarize, the SR algorithm contains only a few hyper-parameters, i.e., sliding window size $\omega$, estimated points number $\kappa$, and anomaly detection threshold $\tau$. We set them empirically and show their robustness in our experiments. Therefore, the SR algorithm is a good choice for online anomaly detection service. 
	
\subsection{SR-CNN}
The original SR method utilizes a single threshold upon the \textit{saliency map} to detect anomaly points, as defined in Equation (\ref{fuc:threshold}). However, this rule is so na\"ive that it is natural to seek for more sophisticated decision rules. Our philosophy is to train a discriminative model on well-designed synthetic data as the anomaly detector. The synthetic data can be generated by injecting anomaly points into a collection of \textit{saliency maps} that are not included in the evaluated data. The injection points are labeled as anomalies while others are labeled as normal. Concretely, we randomly select several points in the time series, calculate the injection value to replace the original point and get its \textit{saliency map}. The values of anomaly points are calculated by:
\begin{align}
x = (\overline{x} + mean)(1 + var) \cdot r + x
\end{align}
where $\overline{x}$ is the local average of the preceding points; $mean$ and $var$ are the mean and variance of all points within the current sliding window; and $r \sim \mathcal{N}(0,1)$ is randomly sampled.
		
We choose CNN as our discrimative model architecture. CNN is a commonly used supervised model for saliency detection~\cite{zhao2015saliency}. However, as we do not have enough labeled data in our scenario, we apply CNN on the basis of \textit{saliency map} instead of raw input, which makes the problem of anomaly annotation to be much easier. In practice, we collect production time-series with synthetic anomalies as training data. The advantage is that the detector can be adaptive to the change of time-series distribution, while no manually labeled data is required. In our experiments, we use totally 65 million points for training.
\begin{figure}[t]
    \centering
	\includegraphics[width=0.45\textwidth]{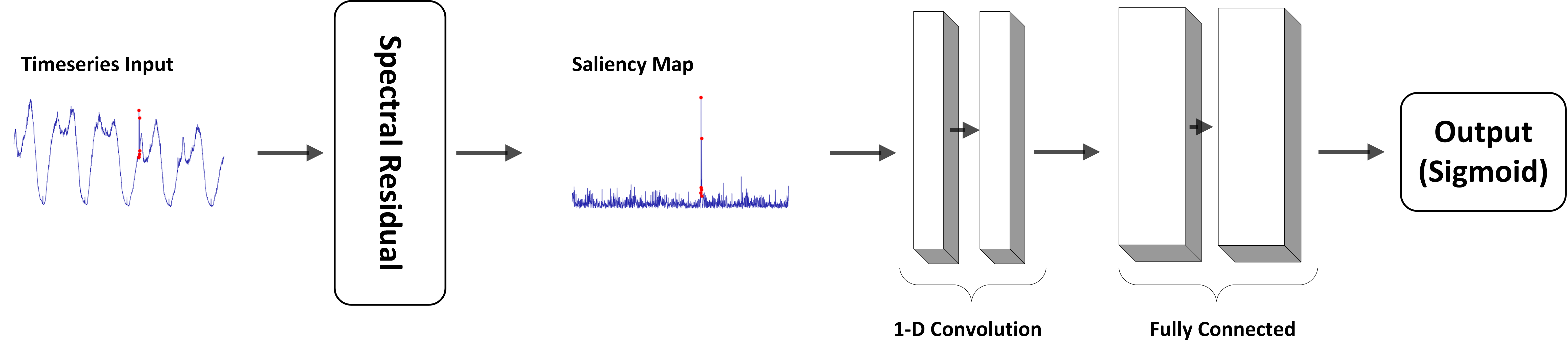}
	\caption{SR-CNN architecture}
	\label{fig:arc}
\end{figure}
The architecture of SR-CNN is visualized in Figure \ref{fig:arc}. The network is composed of two 1-D convolutional layers (with filter size equals to the sliding window size $\omega$) and two fully connected layers. The channel size of the first convolutional layer is equal to $\omega$; while the channel size is doubled in the second convolutional layer. Two full connected layers are stacked before Sigmoid output. Cross entropy is adopted as the loss function; and SGD optimizer is utilized in the training process. 

\section{experiments}
	\subsection{Datasets}
	\label{section:dataset}
	We use three datasets to evaluate our model. KPI and Yahoo are public datasets\footnote{These two datasets are used only for research purpose and do not leveraged in production.} that are commonly used for evaluating the performance of time-series anomaly detection; while Microsoft is an internal dataset collected in the production. These datasets cover time-series of different time intervals and cover a broad spectrum of time-series patterns. In these datasets, anomaly points are labeled as positive samples and normal points are labeled as negative. The statistics of these datasets are shown in Table \ref{Statistics of datasets}.
	
	\textbf{KPI} is released by AIOPS data competition~\cite{kpidataset,kpicomptition}. The dataset consists of multiple KPI curves with anomaly labels collected from various Internet Companies, including Sogou, Tecent, eBay, etc. Most KPI curves have an interval of 1 minute between two adjacent data points, while some of them have an interval of 5 minutes. 
	
	\textbf{Yahoo} is an open data set for anomaly detection released by Yahoo lab\footnote{\url{https://yahooresearch.tumblr.com/post/114590420346/a-benchmark-dataset-for-time-series-anomaly}}. Part of the time-series curves is synthetic (i.e., simulated); while the other part comes from the real traffic of Yahoo services. The anomaly points in the simulated curves are algorithmically generated and those in the real-traffic curves are labeled by editors manually. The interval of all time-series is one hour.
	
	\textbf{Microsoft} is a dataset obtained from our internal anomaly detection service at Microsoft. We select a collection of time-series randomly for evaluation. The selected time-series reflect different KPIs, including revenues, active users, number of pageviews, etc. The anomaly points are labeled by customers or editors manually; and the interval of these time-series is one day.
	\begin{table}
		\renewcommand\arraystretch{1.2}
		\caption{Statistics of datasets}
		\label{Statistics of datasets}
		\begin{threeparttable}
		    \begin{tabularx}{0.48\textwidth}{cccc} \hline
				\textbf{DataSet}& \textbf{Total Curves} & \textbf{Total Points} & \textbf{Anomaly Points}\\ \hline
				\textbf{KPI} & 58& 5922913& 134114/2.26\% \\
				\textbf{Yahoo} & 367& 572966& 3896/0.68\% \\
				\textbf{Microsoft
				} & 372 & 66132 & 1871/2.83\% \\
				\hline
			\end{tabularx}
			\begin{tablenotes}
				\footnotesize
				\item
			\end{tablenotes}
		\end{threeparttable}
	\end{table}
\subsection{Metrics}
\label{section:metrics}
We evaluate our model from three aspects, \textbf{accuracy}, \textbf{efficiency} and \textbf{generality}.
\begin{figure}[t]
    \centering
    \includegraphics[width=0.45\textwidth]{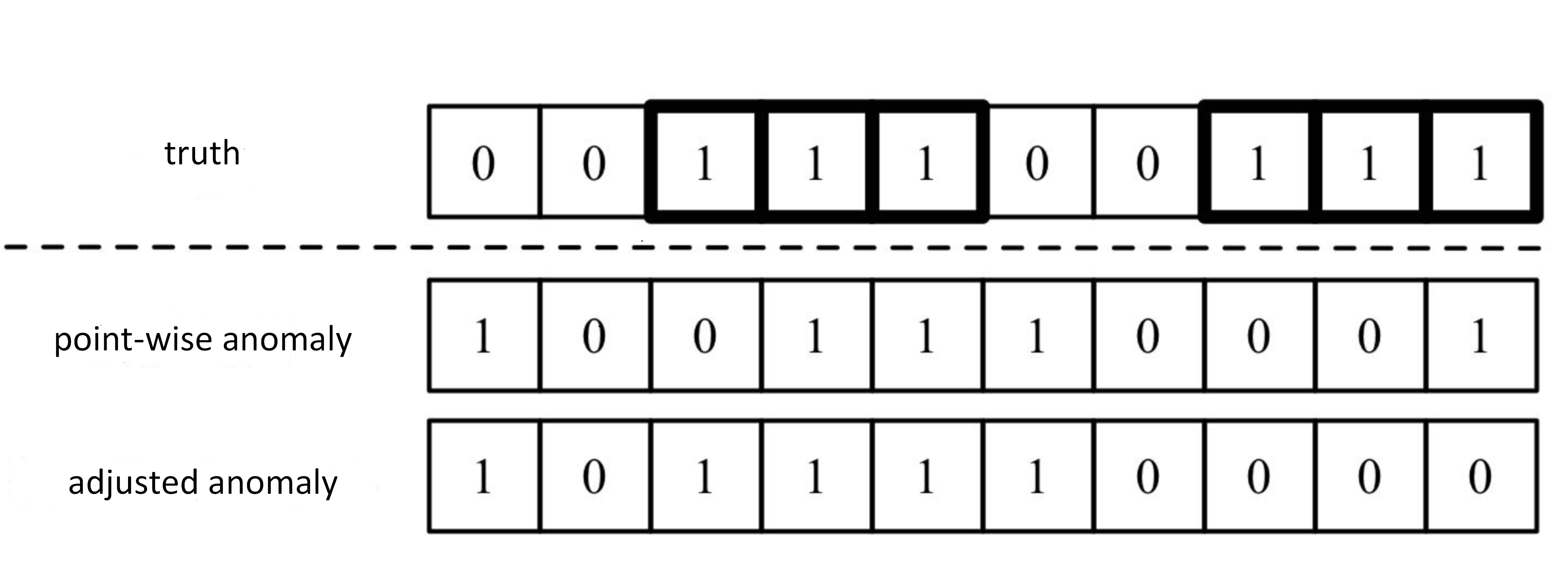}
    \caption{\label{fig:eval_strategy}Illustration of the evaluation strategy. There are 10 contiguous points in the time-series, where the first row indicates ground truth; the second row shows the point-wise anomaly detection results; and the third row shows adjusted results according to the evaluation strategy.}
\end{figure}   
We use precision, recall and $F_{1}$-score to indicate the \textbf{accuracy} of our model. In real applications, the human operators do not care about the point-wise metrics. It is acceptable for an algorithm to trigger an alert for any point in a contiguous anomaly segment if the delay is not too long. Thus, we adopt the evaluation strategy\footnote{The evaluation script is available at https://github.com/iopsai/iops/tree/master/evaluation} following~\cite{xu2018unsupervised}. We mark the whole segment of continuous anomalies as a positive sample which means no matter how many anomalies have been detected in this segment, only one effective detection will be counted. If any point in an anomaly segment can be detected by the algorithm, and the delay of this point is no more than $k$ from the start point of the anomaly segment, we say this segment is detected correctly. Thus, all points in this segment are treated as correct, and the points outside the anomaly segments are treated as normal. 
	
The evaluation strategy is illustrated in Figure \ref{fig:eval_strategy}. As shown in the first row of Figure \ref{fig:eval_strategy}, there are 10 contiguous points and two anomaly segments in the example time-series. The prediction results are shown in the second row. In this case, if we allow the delay as one point, i.e., $k=1$, the first segment is treated as correct and the second is treated as incorrect (because the delay is more than one point). Thus, the adjusted results are illustrated in the third row. Based on the adjusted results, the value of precision, recall and $F_1$-score can be calculated accordingly. In our experiments, we set $k=7$ for minutely time-series, $k=3$ for hourly time-series and $k=1$ for daily time-series following the requirement of real application.

\textbf{Efficiency} is another key indicator of anomaly detection models, especially for those be applied in online services. In the system, we must complete hundreds of thousands of calculations per second. The latency of the model needs to be small enough so that it won't block the whole computation pipeline. In our experiments, we evaluate total execution time on the three datasets to compare the efficiency of different anomaly detection approaches.
	
Besides accuracy and efficiency, we also emphasize \textbf{generality} in our evaluation. As illustrated previously, an industrial anomaly detection model should have the ability to handle different types of time-series. To evaluate generality, we group the time-series in Yahoo dataset into 3 major classes (for example, seasonal, stable and unstable as shown in Figure \ref{fig:series_pattern}) manually and compare the $F_1$-score on different classes separately.
\subsection{SR/SR-CNN Experiment}
\label{section: SR experiment}
    \begin{table*}[ht]
	    \centering
	    \setlength\tabcolsep{4.0pt}
	    \renewcommand\arraystretch{1.2}
		\caption{Result comparison of cold-start}
		\label{table:result}
		\begin{threeparttable}
		 \begin{tabularx}{\textwidth}{l|cccc|cccc|cccc} \hline
                & \multicolumn{4}{c|}{\textbf{KPI}}& \multicolumn{4}{c|}{\textbf{Yahoo}}& \multicolumn{4}{c}{\textbf{Microsoft}}\\\hline
                Model& $F_{1}$-score&$Precision$&$Recall$&$Time(s)$& $F_{1}$-score&$Precision$&$Recall$&$Time(s)$& $F_{1}$-score&$Precision$&$Recall$&$Time(s)$\\\hline
                \textbf{FFT}& \textbf{0.538} & 0.478 & 0.615& 3756.63 & 0.291& 0.202& 0.517& 356.56 & 0.349& 0.812& 0.218& 8.38\\
                \textbf{Twitter-AD}& 0.330& 0.411& 0.276& 523232.0 & 0.245& 0.166& 0.462& 301601.50 & 0.347& 0.716& 0.229& 6698.80\\
                \textbf{Luminol}& 0.417 & 0.306 & 
                0.650 & 14244.92 & \textbf{0.388}& 0.254& 0.818 & 1071.25 & \textbf{0.443}& 0.776 & 0.310& 16.26 \\ \hline
                \textbf{SR}& 0.666 & 0.637 & 0.697& 1427.08 & 0.529& 0.404 & 0.765 & 43.59 & 0.484 & 0.878& 0.334& 2.45\\
                \textbf{SR-CNN}& \textbf{0.732} & 0.811 & 0.667 & 6805.13 &\textbf{0.655} & 0.786&0.561 &279.97 &\textbf{0.537} &0.468 & 0.630& 25.26\\
            \hline
        \end{tabularx}
        \end{threeparttable}
    \end{table*}
    \begin{table*}[!ht]
	    \centering
	    \renewcommand\arraystretch{1.2}
		\caption{Result comparison on test data}
		\label{table:result_training}
		\begin{threeparttable}
		    \begin{tabularx}{\textwidth}{l|cccc|cccc|cccc} \hline
                & \multicolumn{4}{c|}{\textbf{KPI}}& \multicolumn{4}{c|}{\textbf{Yahoo}}& \multicolumn{4}{c}{\textbf{Microsoft}}\\\hline
                Model& $F_{1}$-score & $Precision$ & $Recall$ & $Time(s)$ & $F_{1}$-score & $Precision$ & $Recall$ & $Time(s)$ &$F_{1}$-score & $Precision$ & $Recall$ & $Time(s)$\\\hline
                \textbf{SPOT}& 0.217& 0.786& 0.126& 9097.85 & \textbf{0.338}& 0.269& 0.454& 2893.08& 0.244& 0.702& 0.147& 9.43\\
                \textbf{DSPOT}& \textbf{0.521}& 0.623& 0.447& 1634.41 & 0.316& 0.241& 0.458& 339.62 & 0.190& 0.394& 0.125& 1.37 \\
                \textbf{DONUT}& 0.347&0.371& 0.326& 24248.13 & 0.026& 0.013& 0.825& 2572.76 & \textbf{0.323}& 0.241& 0.490& 288.36\\\hline
                \textbf{SR}& 0.622 & 0.647 & 0.598 & 724.02 & 0.563& 0.451& 0.747& 22.71 & 0.440& 0.814& 0.301& 1.55 \\
                 \textbf{SR-CNN}& \textbf{0.771} & 0.797 & 0.747 & 2724.33 & \textbf{0.652} & 0.816 & 0.542 & 125.37 &\textbf{0.507} &0.441 & 0.595&16.13\\
                \hline
        \end{tabularx}
        \end{threeparttable}
    \end{table*}
    \begin{table}
		\renewcommand\arraystretch{1.2}
		\caption{Generality Comparison on Yahoo dataset}
		\label{table:generality}
		\begin{threeparttable}
		    \begin{tabularx}{0.48\textwidth}{ccccccccc} \hline
				& Seasonal & Stable & Unstable & Overall & $Var$\\ \hline
				\textbf{FFT} & \textbf{0.446}	& 0.370 & 0.301 & 0.364 & \textbf{0.060}  \\
				\textbf{Twitter-AD} & 0.397 & \textbf{0.924} & \textbf{0.438} & \textbf{0.466} & 0.268 \\
				\textbf{Luminol} & 0.374 & 0.763 & 0.428 & 0.430 & 0.195  \\
				\textbf{SPOT} & 0.199 & 0.879 & 0.356 & 0.338 & 0.322 \\
				\textbf{DSPOT} & 0.211 & 0.485 &	0.379 &	0.316 &	0.120  \\
				\textbf{DONUT} & 0.023 &	0.032 &	0.029 & 0.026 &	0.004 \\\hline
				\textbf{SR} & 0.558 & 0.601 & \textbf{0.556} & 0.563 & \textbf{0.023} \\
				\textbf{SR-CNN} & \textbf{0.716} & \textbf{0.752} & 0.464 & \textbf{0.652} & 0.128 \\
				\hline
			\end{tabularx}
			\begin{tablenotes}
				\footnotesize \item {$Var$ indicates the standard deviation of the overall $F_{1}$-scores for the three classes}
			\end{tablenotes}
		\end{threeparttable}
	\end{table}
    We compare SR and SR-CNN with state-of-the-art unsupervised time-series anomaly detection methods. The baseline models include FFT (Fast Fourier Transform)~\cite{rasheed2009fourier},  Twitter-AD (Twitter Anomaly Detection)~\cite{twitter-ad}, Luminol (LinkedIn Anomaly Detection)~\cite{linkedin/luminol}, DONUT~\cite{xu2018unsupervised}, SPOT and DSPOT~\cite{siffer2017anomaly}. Among these methods, FFT, Twitter-AD and Luminol do not need additional data to start, so we compare these models in a cold-start setting by treating all the time-series as test data. On the other hand, SPOT, DSPOT and DONUT need additional data to train their models. Therefore, we split the points of each time-series as two halves according to the time order. The first half is utilized for training those unsupervised models while the second half is leveraged for evaluation. Note that DONUT can leverage additional labeled data to benefit the anomaly detection performance. However, as we are aiming to get a fair comparison in the fully unsupervised setting, we do not use additional labeled data in the implementation\footnote{https://github.com/haowen-xu/donut}. 
    
    The experiments are conducted in a streaming pipeline. The points of a time-series are ingested to the evaluation pipeline sequentially. In each turn, we only detect if the recent point is anomaly or not while the succeeding points are invisible. In the setting of cold-start, recommended configurations are applied to the baseline models which come from papers or codes published by the authors. For SR and SR-CNN, we set the hyper-parameters empirically. In SR, shape of $h_q(f)$ q is set as 3, number of local average of preceding points z is set as 21, threshold $\tau$ is set as 3, number of estimated points $\kappa$ is set as 5, and the sliding window size $\omega$ is set as 1440 on KPI, 64 on Yahoo and 30 on Microsoft. For SR-CNN, q, z, $\kappa$ and $\omega$ are set to the same value.
    
    We report (1) $F_{1}$-score; (2) $Precision$; (3) $Recall$;  and (4) CPU execution times separately for each dataset. We can see that SR significantly outperforms current state-of-the-art unsupervised models. Furthermore, SR-CNN achieves further improvement on all three datasets, which shows the advantage of replacing the single threshold by a CNN discriminator. Table \ref{table:result} shows comparison results of FFT, Twitter-AD and Luminol in the cold-start scenario. We improve the $F_1$-score by 36.1\% on KPI dataset, 68.8\% on Yahoo dataset and 21.2\% on Microsoft dataset compared to the best results achieved by baseline solutions. Table \ref{table:result_training} demonstrates the comparison results of those unsupervised models which need to be trained on the first half of the dataset (labels are excluded). As shown in Table \ref{table:result_training}, the $F_1$-score is improved by 48.0\% on KPI dataset, 92.9\% on Yahoo dataset and 57.0\% on Microsoft dataset than the best state-of-the-art results. 
    
    Moreover, SR is the most efficient method as indicated by the total CPU execution time in Table \ref{table:result} and \ref{table:result_training}. SR-CNN achieves better accuracy with a reasonable latency increase. For generality comparison, we conduct the experiments on the second half of Yahoo dataset, which is classified into three classes manually. $F_{1}$-score on different classes of Yahoo dataset is reported separately in Table \ref{table:generality}. SR and SR-CNN achieve outstanding results on various patterns of time-series. SR is the most stable one across the three classes. SR-CNN also demonstrates good  capability of generalization.
	
	\subsection{SR+DNN}
	\label{section:supervised}
	     \begin{table*}[t]
		\centering
		\renewcommand\arraystretch{1.2}
		\caption{Features used in the supervised DNN model}
		\label{features}
		\begin{threeparttable}
			\begin{tabularx}{1\textwidth}{|l|X|} \hline
			    Feature & Description \\ \hline
				 Transformations & Transformations to the value of each data point. We use logarithm as our transformation function and leverage the result value as a feature. \\ \hline
                 Statistics & We applied sliding windows to the time-series and treat the statistics calculated in each sliding window as features. The statistics we used include mean, exponential weighted mean, min, max, standard deviation, and the quantity of the data point values within a sliding window. We use multiple sizes of the sliding window to generate different features. The sizes are [10, 50, 100, 200, 500, 1440] \\ \hline
                Ratios & The ratios of current point value against other statistics or transformations\\ \hline
                Differences & The differences of current point value against other statistics or transformations \\ \hline
			\end{tabularx}
		\end{threeparttable}
	\end{table*}
	In the previous experiments, we can see that the SR model shows convincing results in the unsupervised anomaly detection scenario. However, when labels of anomalies are available, we can obtain more satisfactory results as illustrated in previous works~\cite{liu2015opprentice}. Thus, we would like to know whether our methodology contributes to the supervised scenario as well. Concretely, we treat the intermediate results of SR as an additional feature in the supervised anomaly detection model. We conduct the experiment on KPI dataset as it has been extensively studied in the AIOPS data competition~\cite{kpicomptition}.
	
\begin{figure}[t]
		\centering 
		\includegraphics[width=0.45\textwidth]{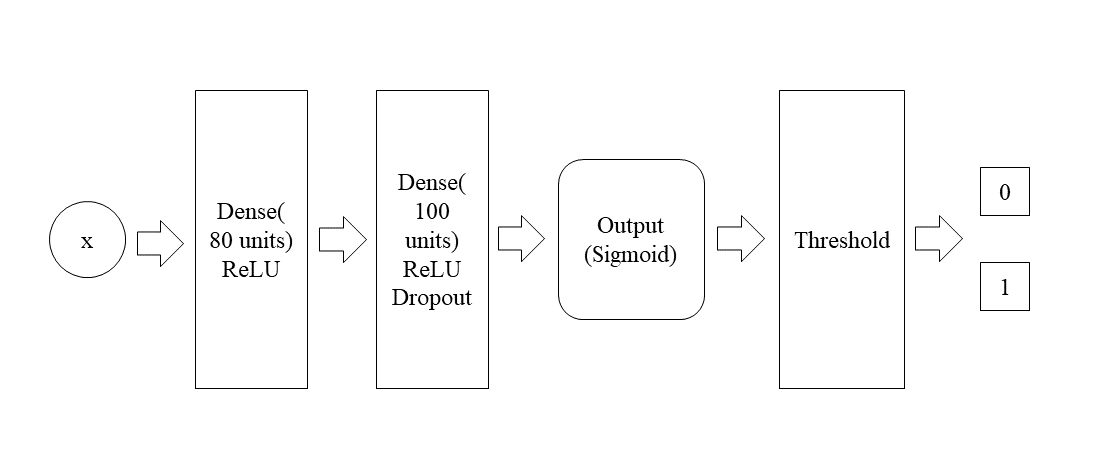}
		\caption{DNN architecture}
		\label{fig:supervise}
	\end{figure}

	We adopt the DNN-based supervised model~\cite{dnn-champion} which is the champion in the AIOPS data competition. The DNN architecture is composed by an input layer, an output layer and two hidden layers (shown in Figure \ref{fig:supervise}). We add a dropout layer after the second hidden layer and set dropout ratio as 0.5. In addition, we apply $L_1=L_2=0.0001$ regularization to the weights of all layers. Since the output of the model indicates the likelihood of a data point being an anomaly, we search for the optimal threshold on the training set.
	
	Each data point is associated with a feature vector, which consists of different types of features including transformations, statistics, ratios, and differences (Table \ref{features}). We follow the official train/test split of the dataset, where the statistics is shown in Table \ref{train and test}. We can see that the proportion of positive and negative samples is extremely imbalanced. Thus, we train our model by over-sampling anomalies to keep the positive/negative proportion to 1:2.
    
\begin{figure}[t]
	\centering 
	\includegraphics[width=0.35\textwidth, height=0.25\textwidth]{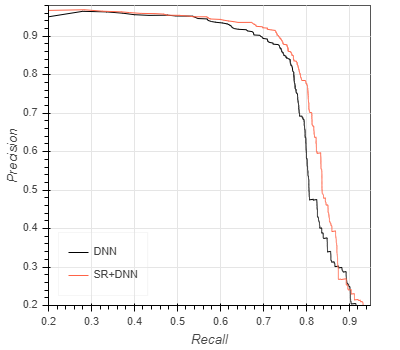}
	\caption{P-R curves of SR+DNN and DNN methods}
	\label{fig:pr_curve}
	\end{figure}

    Experimental results are shown in Table \ref{Supervised results on KPI dataset}. We can see that the SR feature brings 1.6\% improvement in $F_1$-score to the vanilla DNN model. Especially, the SR-powered DNN model establishes a new state-of-the-art on the KPI dataset. To the best of our knowledge, it is the best-ever result reported on the KPI dataset up to the date of paper submission. Moreover, we draw the P-R curve of the SR+DNN and DNN methods. As illustrated in Figure \ref{fig:pr_curve}, SR+DNN outperforms the vanilla DNN consistently on various threshold. 
    \begin{table}[t]
		\centering
		\renewcommand\arraystretch{1.2}
		\caption{Train and test split of KPI dataset}
		\label{train and test}
		\begin{threeparttable}
			\begin{tabular}{l|c|c} \hline
				\textbf{DataSet} & \textbf{Total points} & \textbf{Anomaly points} \\ \hline
				Train & 3004066 & 79554/2.65\% \\ \hline
				Test & 2918847 & 54560/1.87\% \\ \hline
			\end{tabular}
		\end{threeparttable}
	\end{table}
	
	\begin{table}
		\centering
		\renewcommand\arraystretch{1.2}
		\caption{Supervised results on KPI dataset}
		\label{Supervised results on KPI dataset}
		\begin{threeparttable}
			\begin{tabular}{l|c|c|c} \hline
				\textbf{Model} & \textbf{$F_{1}$-score} & \textbf{Precision} & \textbf{Recall} \\ \hline
				DNN & 0.798& 0.849& 0.753\\ \hline
				SR+DNN & \textbf{0.811}& 0.915& 0.728 \\
				\hline
			\end{tabular}
		\end{threeparttable}
	\end{table}
	
\section{related works}
	\subsection{Anomaly detectors}
	Previous works can be categorized into statistical, supervised and unsupervised approaches. In the past years, several models were subsequently proposed in the statistics literature, including hypothesis testing~\cite{rosner1983percentage}, wavelet analysis~\cite{lu2009network}, SVD~\cite{mahimkar2011rapid} and auto-regressive integrated moving average (ARIMA)~\cite{zhang2005network}. Fast Fourier Transform (FFT)~\cite{van1992computational} is another traditional method for time-series processing. For example,~\cite{rasheed2009fourier} highlighted the areas with high frequency change by FFT and reconfirmed it with Z-value test. In 2015, Twitter~\cite{twitter-ad} proposed a model to detect anomalies in time-series of both application metrics (e.g., Tweets Per Sec) and system metrics (e.g., CPU utilization). In 2017, SPOT and DSPOT~\cite{siffer2017anomaly} were proposed on the basis of Extreme Value Theory~\cite{de2007extreme}, the threshold of which can be selected automatically. 
	
	The performances of traditional statistical models are not satisfactory in real applications. Thus, researchers have investigated supervised models to improve the anomaly detection accuracy. 
	Opprentice~\cite{liu2015opprentice} outperformed other traditional detectors by using statistical detectors as feature extractors and leveraged a Random Forest classifier~\cite{liaw2002classification} to detect anomalies. Yahoo EGADS~\cite{yahooo2015EGADS} utilized a collection of anomaly detection and forecasting models with an anomaly filtering layer for scalable anomaly detection on time-series data. In 2017, Google leveraged deep learning models to detect anomalies on their own dataset~\cite{anomalyGoogle} and achieved promising results. However, continuous labels can not be obtained in industrial environment, which makes these supervised approaches insufficient in online applications.
	
	As a result, advanced unsupervised approaches have been studied to tackle the problem in industrial application. In 2018, \cite{xu2018unsupervised} proposed DONUT, an unsupervised anomaly detection method based on Variational Auto-Encoder (VAE)~\cite{doersch2016tutorial}. VAE was leveraged to model the reconstruction probabilities of normal time-series, while the abnormal points were reported if the reconstruction error was larger than a threshold. Besides, LinkedIn developed Luminol \cite{linkedin/luminol} based on \cite{linkedinref}, which segmented time-series into chunks and used the frequency of similar chunks to calculate anomaly scores.

	\subsection{Saliency detection approaches}
	\label{Saliency detection}
	Our work has been inspired by visual saliency detection models. Hou et al.~\cite{hou2007saliency} invented the Spectral Residual (SR) model for saliency detection and demonstrated impressive performance in their experiments. They assumed that an image can be divided into redundant part and innovation part, while people's vision is more sensitive to the innovation part. Meanwhile, the log amplitude spectrum of an image subtracting the average log amplitude spectrum captures the saliency part of the image. Guo et al.~\cite{guo2008spatio} argued that only phase spectrum was enough to detect the saliency part of an image and simplified the algorithm in \cite{hou2007saliency}. Hou et al.~\cite{hou2012image} also proposed an image signature approach for highlighting sparse salient regions with theoretical proof. Although the latter two solutions showed improvement in their publications, we found that Spectral Residual (SR) was more effective in our time-series anomaly detection scenario. Moreover, supervised models based on neural networks are also used in saliency detection. For instance, Zhao et al.~\cite{zhao2015saliency} tackled the problem of salient object detection by a multi-context deep learning framework based on CNN architecture.
	
\section{Conclusion \& Future Work}
Time-series anomaly detection is a critical module to ensure the quality of online services. An efficient, general and accurate anomaly detection system is indispensable in real applications. In this paper, we have introduced a time-series anomaly detection service at Microsoft. The service has been used by more than 200 teams within Microsoft, including Bing, Office and Azure. Anomalies are detected from 4 million time-series per minute maximally in the production. Moreover, we for the first time apply the Spectral Residual (SR) model in the time-series anomaly detection task and innovatively combine the SR and CNN model to achieve an outstanding performance. In the future, we plan to ensemble the state-of-the-art methods together to provide a more robust anomaly detection service to our customers. Besides internal serving, our time-series anomaly detection service will be published on Microsoft Azure as part of Cognitive Service\footnote{https://azure.microsoft.com/en-us/services/cognitive-services/} shortly to external customers.

\bibliographystyle{ACM-Reference-Format}
\bibliography{sample-base}

\end{document}